\documentclass[letterpaper, 10 pt, conference]{ieeeconf}  

\IEEEoverridecommandlockouts                              

\usepackage{multirow}
\usepackage{lipsum}
\usepackage{amsmath}
\usepackage{amssymb}
\usepackage{siunitx}
\usepackage{bm}
\usepackage{graphicx}
\usepackage{subcaption}
\usepackage{booktabs}
\usepackage{comment}
\usepackage[acronym]{glossaries}
\makeglossaries
\newacronym{adas}{ADAS}{Advanced Driver Assistance Systems}
\newacronym{gnss}{GNSS}{Global Navigation Satellite System}
\newacronym{rf}{RF}{Radio Frequency}
\newacronym{nig}{NIG}{Normal-Inverse-Gamma}
\newacronym{ssm}{SSM}{State Space Model}
\newacronym{evimamba}{EVC-Mamba}{Evidential Velocity Correction using Mamba}
\newacronym{nll}{NLL}{Negative Log Likelihood}
\newacronym{bf}{BF}{Body Frame}
\newacronym{hf}{HF}{Horizontal Frame}
\newacronym{es}{ES}{Error State}
\newacronym{ins}{INS}{Inertial Navigation System}

\newacronym{piml2}{PRML2}{Physics-Regularized Machine Learning for Localization}
\newacronym{kf}{KF}{Kalman Filter}
\newacronym{ekf}{EKF}{Extended Kalman Filter}
\newacronym{rmse}{RMSE}{Root Mean Squared Error}
\newacronym{gru}{GRU}{Gated Recurrent Unit}
\newacronym{aeb}{AEB}{Autonomous Emergency Braking}
\newacronym{esc}{ESC}{Electronic Stability Control}
\newacronym{mems}{MEMS}{Micro-Electro-Mechanical System}
\newacronym{vsa}{VSA}{Vehicle Sideslip Angle}
\newacronym{ml}{ML}{Machine Learning}
\newacronym{ai}{AI}{Artificial Intelligence}
\newacronym{vmm1}{VMM-1}{Vehicle Motion Model - 1}
\newacronym{vmm2}{VMM-2}{Vehicle Motion Model - 2}
\newacronym{vmm}{VMM}{Vehicle Motion Model}
\newacronym{vmms}{VMMs}{Vehicle Motion Models}
\newacronym{esekf}{ES-EKF}{Error-State Extended Kalman Filter}
\newacronym{uahi}{UAHL}{Uncertainty-Aware Hybrid Learning}
\newacronym{obd}{OBD}{On Board Diagnostics}
\newacronym{ros}{ROS}{Robot Operating System}
\newacronym{can}{CAN}{Controller Area Network}
\newacronym{pid}{PID}{Parameter IDentification}
\newacronym{gps}{GPS}{Global Positioning Systems}
\newacronym{lstm}{LSTM}{Long Short-Term Memory}
\newacronym{cg}{COG}{Centre of Gravity}
\newacronym{gmm}{GMM}{Gaussian Mixture Model}
\newacronym{revsted}{ReV-StED}{Real-world Vehicle State Estimation Dataset}
\newacronym{sota}{SOTA}{State-of-the-art}
\newacronym{imu}{IMU}{Inertial Measurement Unit}
\newacronym{dgnss}{DGNSS}{Differential Global Navigation Satellite System}
\newacronym{irs}{IRS}{Inertial Reference System}
\newacronym{adma}{ADMA}{Automotive Dynamic Motion Analyzer}
\newacronym{me}{MaxE}{Max Error}
\newacronym{mse}{MSE}{Mean Squared Error}
\newacronym{mae}{MAE}{Mean Absolute Error}
\newacronym{ptp}{PTP}{Precision Time Protocol}
\newacronym{poi}{PoI}{Point of Installation}
\newacronym{cog}{CoG}{Center of Gravity}
\newacronym{gmr}{GMR}{Gaussian Mixture Regression}
\newacronym{osd}{OSD}{Onboard Sensor Data}
\usepackage[ruled,vlined]{algorithm2e}
\usepackage{graphicx}
\graphicspath{{./Figures/}}

\def\eqref#1{equation~\ref{#1}}

\def\1{\bm{1}}

\def\va{{\bm{a}}}
\def\vb{{\bm{b}}}

\def\vg{{\bm{g}}}
\def\vh{{\bm{h}}}

\def\vo{{\bm{o}}}
\def\vp{{\bm{p}}}

\def\vu{{\bm{u}}}
\def\vv{{\bm{v}}}

\def\vx{{\bm{x}}}

\def\mO{{\bm{O}}}

\def\mR{{\bm{R}}}

\def\mT{{\bm{T}}}

\DeclareMathAlphabet{\mathsfit}{\encodingdefault}{\sfdefault}{m}{sl}
\SetMathAlphabet{\mathsfit}{bold}{\encodingdefault}{\sfdefault}{bx}{n}

\usepackage[english]{babel}
\usepackage[hyphens]{url}
\usepackage[hidelinks]{hyperref}
\title{
\LARGE \bf Uncertainty-Aware Velocity Correction for Proprioceptive Vehicle Localization using Evidential Mamba}

\author{
	Abinav Kalyanasundaram$^{1}$, Karthikeyan Chandra Sekaran$^{1}$, Wolfgang Utschick$^{3}$ and  Michael Botsch$^{1}$
    \thanks{$^{1}$AImotion Bavaria, Technische Hochschule Ingolstadt, Germany, {\tt\small firstname.lastname@thi.de}}%
    \thanks{$^{3}$Technische Universität München, Germany, {\tt\small utschick@tum.de}}%
}

\begin{document}
\bstctlcite{IEEEexample:BSTcontrol}
\maketitle
\thispagestyle{empty}
\pagestyle{empty}

\begin{abstract}
Reliable localization in GNSS-denied environments remains a fundamental challenge for intelligent vehicles, as inertial navigation systems accumulate unbounded drift without external correction. 
Existing approaches provide drift correction through dedicated infrastructure, expensive external sensors, or complex multi-sensor fusion, each introducing practical deployment barriers.  
We propose \gls{evimamba}, a learning-based architecture that transforms onboard vehicle sensor data into a virtual velocity sensor for IMU drift correction without additional hardware.
A Mamba-based selective state space model captures the temporal dynamics of vehicle motion, while evidential deep learning with a Normal-Inverse-Gamma distribution provides principled uncertainty quantification. 
The resulting uncertainty-aware velocity estimate is incorporated as a virtual correction measurement into an Error-State Extended Kalman Filter to reduce position drift. 
Evaluation on real-world vehicle data demonstrates that inertial navigation using the proposed velocity correction achieves localization accuracy within 10\% of a dedicated external velocity sensor across different outage durations. 
The proposed architecture supports real-time onboard deployment at 40~Hz on edge hardware, enabling reliable localization during prolonged GNSS outages.
\end{abstract}

\section{Introduction}
Accurate localization is a key requirement for the safe and reliable operation of intelligent vehicles, enabling essential functions such as path planning, motion control, and automated driving~\cite{intro_localization}.
In general, precise localization is achieved by fusing an~\gls{ins} with~\gls{gnss} signals to compensate for drift~\cite{gnss}. 
However, in \gls{gnss}-denied indoor environments, such as parking structures and tunnels, position drift grows unbounded due to the absence of external corrections~\cite{lstmIMU}.

Localization methods for \gls{gnss}-denied environments are broadly classified into \gls{rf}-based, vision-based, multi-sensor fusion, and proprioceptive inertial navigation approaches~\cite{indoor_survey}. 
\gls{rf}-based techniques depend on dedicated infrastructure and suffer from signal attenuation and multipath effects~\cite{rf-based}.
Vision-based methods using cameras or LiDARs can achieve high accuracy, yet remain sensitive to illumination changes and occlusions~\cite{localisation_survey}. 
Multi-sensor fusion improves robustness at the cost of increased system complexity and expense~\cite{vio_eccv,vins}. 
In contrast, \gls{imu}-based inertial navigation offers a low-cost, infrastructure-free solution but suffers from long-term drift due to noise and bias accumulation~\cite{noise_imu}. 
Recent data-driven methods have improved \gls{imu}-based dead reckoning~\cite{lstmIMU,noise_imu}, but their effectiveness remains limited to shorter outage durations~\cite{imu_survey}. 
For longer outages, additional hardware, such as Correvit sensors~\cite{sven}, is typically required, which can be costly~\cite{wheel-imu}.
Alternatively, readily available onboard vehicle sensor data, such as steering wheel angle and wheel speeds, can be leveraged to construct a virtual velocity sensor, analogous to vehicle sideslip angle estimation~\cite{abinav}.

\begin{figure}[tb]
    \centering
        \includegraphics[scale=1.25, trim=10pt 10pt 10pt 10pt, clip]{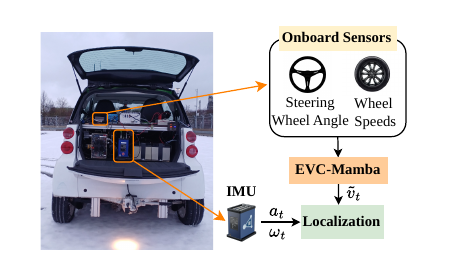}
    \caption{Overview of the test vehicle configuration and the proposed \gls{evimamba} based velocity correction for proprioceptive localization using \gls{imu}.}
    \label{fig:piml_architecture}
    \vspace{-0.5cm}
\end{figure} 

Building on this insight, this work introduces \textbf{E}vidential \textbf{V}elocity \textbf{C}orrection using \textbf{Mamba}~(\gls{evimamba}), a learning-based 
architecture that leverages onboard vehicle sensors for precise velocity correction in IMU-based localization, as illustrated in Fig.~\ref{fig:piml_architecture}.
Our method employs a Mamba-based selective \gls{ssm}~\cite{mamba} for velocity estimation and models uncertainty via evidential deep learning with a \gls{nig} distribution. 
The uncertainty-aware velocity estimate is then fused as a virtual correction measurement within an \gls{esekf}, improving proprioceptive localization 
in GNSS-denied environments.
The main contributions of this work are as follows:

\begin{itemize}
\item \textbf{\gls{gnss}-Independent Velocity Correction:} A learning-based virtual velocity sensor that exploits onboard vehicle sensors to correct \gls{imu} drift, without requiring any external sensing hardware or infrastructure.
\item \textbf{Uncertainty-Aware Velocity Estimation:} The proposed architecture leverages a Mamba-\gls{ssm} to model vehicle motion and jointly predicts velocity with its uncertainty using an evidential deep learning framework, enabling robust and reliable correction signals.
\item \textbf{\gls{esekf} Integration:} The \gls{evimamba} predictions are incorporated as a virtual measurement within an \gls{esekf} alongside \gls{imu} mechanization.
\item \textbf{Real-time Capability:} Achieves accuracy comparable to external velocity sensors on real-world vehicle data, while remaining suitable for onboard deployment.\footnote{The code to replicate results will be released upon acceptance.}
\end{itemize}



\section{Related works}
Proprioceptive localization for ground vehicles is conventionally addressed using recursive Bayesian filtering methods such as the \gls{kf}, \gls{ekf}, Error-State \gls{ekf}, and particle filters~\cite{survey_localization}. 
These approaches fuse motion models with \gls{imu} measurements to estimate vehicle states, but suffer from drift due to bias and noise accumulation, leading to quadratic growth of position error over time~\cite{lstmIMU}.
To mitigate this, precise velocity measurements from external sensors or wheel-mounted \gls{imu}s have been incorporated in~\cite{wheel-imu}, reducing drift to linear growth.
Alternatively, vehicle velocities can be inferred from onboard sensor data such as wheel speeds and steering angles using kinematic models~\cite{onboardsensors-velocity}. 
However, such approaches rely on simplifying assumptions that are often violated in real-world driving conditions, limiting their robustness~\cite{icra_wheelodometry}. 
These limitations motivate learning-based approaches that improve inertial navigation and sensor modeling by learning relationships directly from data~\cite{noise_imu}.
CalibNet~\cite{calibnet2022} demonstrated that deep learning can enhance low-grade IMU performance to mimic that of high-grade sensors.
Recent works~\cite{abinav,TUdelft} show that \gls{ml} models can transform onboard measurements into accurate vehicle dynamic state estimates without complex vehicle models.

Uncertainty-aware learning, particularly evidential deep learning, enables joint prediction and uncertainty quantification by parameterizing higher-order distributions~\cite{evidential_regression}, and has shown improved robustness in perception and mapping tasks~\cite{capellier2021fusion}. 
However, its application to proprioceptive velocity estimation and integration into recursive state estimation remains underexplored.
Modeling temporal dependencies in onboard sensor data is critical for such learning-based approaches.
Sequential modeling in inertial odometry has traditionally relied on recurrent networks (e.g., IONet)~\cite{ionet}, despite Transformers offering improved long-range reasoning at higher computational cost~\cite{transfuser}. 
Structured State Space Models (SSMs), such as S4 and Mamba~\cite{mamba}, provide efficient linear-time alternatives to Transformers but remain largely unexplored in vehicle state estimation. Mamba's selective state space mechanism adapts to input-dependent dynamics, making it well-suited for vehicle sensor data where motion characteristics vary with driving conditions.


To the best of our knowledge, no existing approach jointly addresses learning-based velocity estimation from onboard vehicle sensors, principled uncertainty quantification, and efficient sequence modeling. 
We propose \gls{evimamba}, which combines a Mamba-based virtual velocity sensor with evidential uncertainty estimation and integrates it into an \gls{esekf}, enabling real-time velocity correction for inertial navigation.

\section{Methodology}
This section presents the problem formulation, the
preliminaries, and the proposed \gls{evimamba} architecture. 
The overall framework is illustrated in Fig.~\ref{fig:evlmamba}.

\subsection{Problem Formulation}
This work addresses vehicle localization using \gls{imu} and onboard vehicle sensors in \gls{gnss}-denied conditions. The objective is to estimate the dead-reckoning pose $\mT_t \in SE(3)$ at each time step $t$, where $\mT_t$ represents a 3D rigid-body transformation composed of vehicle’s rotation \mbox{$\mR_t \in SO(3)$} and position \mbox{$\vp_t \in \mathbb{R}^3$} in a global coordinate frame. The pose $\mT_t$ is typically estimated using \gls{imu} measurements $\vu_t = [\va_t, \boldsymbol{\omega}_t]$ within a recursive Bayesian filter to track the vehicle state given by 
\begin{equation}
\label{Eq:function_transformation}
    \vx_{t} = \left[\vp_t, \boldsymbol{\theta}_t, \vv_t \right]^{T}.
\end{equation}
Here, $\vv_t$ denotes the vehicle's linear velocity. The rotation $\mR_t$ is parameterized by the vehicle's orientation $\boldsymbol{\theta}_t$.

This work proposes an uncertainty-aware mapping function $\mathcal{G}_{\boldsymbol{\Theta}}$ to transform \gls{osd} into a precise velocity estimate $\tilde{\vv}_t$ with its associated 
uncertainty $\delta_{\tilde{\vv}}$, formulated as an evidential regression model:
\begin{equation}
\label{Eq:g_theta}
(\tilde{\vv}_t, \delta_{\tilde{{\vv}}}) = \mathcal{G}_{\boldsymbol{\Theta}}(\mO_t; L, m)
\end{equation}
where \mbox{$\mO_t = \{\vo_{t-L+1}, \dots, \vo_t\} \in \mathbb{R}^{L \times m}$} denotes the \gls{osd} for an observation window of $L$ time steps. 
Each measurement vector \ensuremath{\vo_t = [o_t^1, \dots, o_t^m]^\top \in \mathbb{R}^m} contains the readings from $m$ onboard sensors at time step $t$. 
The estimated velocity and uncertainty act as a correction signal within an \gls{esekf} to improve IMU-based pose estimation. A \gls{gnss}/RTK-aided \gls{ins} provides ground truth reference for vehicle states.

\subsection{Preliminaries}
\label{sec:Preliminaries}
Transformer-based architectures, while effective for sequential modeling, incur quadratic complexity with respect to sequence length due to 
self-attention, limiting their suitability for real-time deployment~\cite{mamba}. Mamba
addresses this by introducing a selective \gls{ssm} that achieves linear-time inference, mapping an input sequence $x_t \in \mathbb{R}$ 
to an output $y_t \in \mathbb{R}$ through a latent state $\vh_t \in \mathbb{R}^{N}$ via the discrete-time recurrence:
\begin{align}
    \vh_t &= \mathbf{A}\vh_{t-1} + \mathbf{B}x_t \;, \quad y_t = \mathbf{C}\vh_t
\end{align}
where $\mathbf{A} \in \mathbb{R}^{N \times N}$ and 
$\mathbf{B} \in \mathbb{R}^{N \times 1}$ are discretized 
from their continuous-time counterparts via the timescale 
$\Delta \in \mathbb{R}^+$, and $\mathbf{C} \in \mathbb{R}^{1 \times N}$ 
is the output projection matrix.
Unlike conventional \glspl{ssm} with fixed parameters, Mamba's selective mechanism~\cite{mamba} makes $\mathbf{B}$, $\mathbf{C}$, and $\Delta$ input-dependent:
\begin{align}
    \mathbf{B}_t = \phi_\mathbf{B}(x_t), \quad 
    \mathbf{C}_t = \phi_\mathbf{C}(x_t), \quad 
    \Delta_t = \phi_\Delta(x_t)
\end{align}
where $\phi_\mathbf{B}$, $\phi_\mathbf{C}$, and $\phi_\Delta$ are learned linear projections. Given an input $\mathbf{U} \in \mathbb{R}^{L \times d}$, 
a single Mamba block applies this selective \gls{ssm} independently across the $d$ feature channels and along the temporal dimension $L$, producing a contextualized output of the same shape $\mathbb{R}^{L \times d}$.
\begin{figure*}[htbp]
\centering
\includegraphics[scale=0.90, trim=20pt 5pt 20pt 5pt, clip]{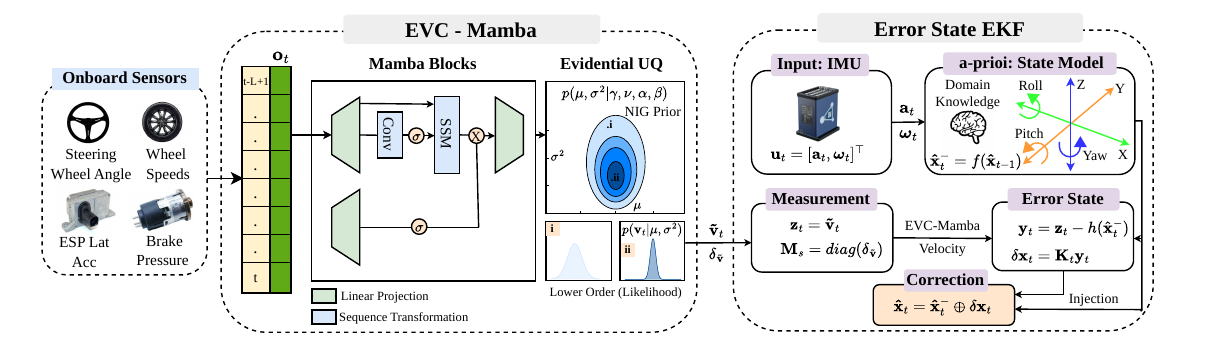}
\caption{Overview of the proposed \gls{evimamba} framework 
for inertial navigation in \gls{gnss}-denied environments. 
Onboard sensor measurements $\mO_t$ are processed through 
a Mamba-based selective \gls{ssm} with an evidential output 
head, which predicts the \gls{nig} parameters 
$m = (\gamma, \nu, \alpha, \beta)$ over the true velocity 
distribution. The \gls{nig} prior yields a lower-order Gaussian likelihood, from which the virtual velocity estimate 
$\tilde{\vv}_t$ and its uncertainty $\delta_{\tilde{\vv}}$ 
are extracted. These estimates are incorporated as a virtual 
measurement into an \gls{esekf} to reduce position 
drift during prolonged GNSS outages.}
\label{fig:evlmamba} 
\vspace{-0.3cm}
\end{figure*}

\subsection{\gls{evimamba} Architecture}
\label{sec:evimamba model}
The proposed~\gls{evimamba} architecture learns the mapping function $\mathcal{G}_{\boldsymbol{\Theta}}$ in Eq.~(\ref{Eq:g_theta}) by combining a selective \gls{ssm} model with evidential velocity regression, as illustrated in Fig.~\ref{fig:evlmamba}.
The Mamba-based sequence model~\cite{mamba} enables temporal modeling of onboard sensor data $\mO_t$ for precise velocity $\tilde{\vv}_t$ estimation, while evidential learning provides uncertainty-aware estimates.

\subsubsection{\textbf{Mamba-based Velocity Estimation}}
Accurate velocity estimation requires modeling of both short-term dynamics and long-range temporal dependencies arising from vehicle motion and sensor noise characteristics~\cite{abinav}. 
This estimation task is formulated as a time-series regression problem, where a sequence of $L$ onboard sensor measurements $\mO_t \in \mathbb{R}^{L \times m}$ is used to infer the vehicle velocity. We employ a Mamba-\gls{ssm}~\cite{mamba} for efficient 
linear-time temporal modeling of the vehicle sensor data.

Each onboard measurement in the input sequence $\mO_t$, denoted as $\vo_t \in \mathbb{R}^m$, is projected into a $d$-dimensional latent space via an up-projection layer $\phi_\text{u}$, producing input embeddings $\mathbf{U}_{\mathrm{enc}} \in \mathbb{R}^{L \times d}$. These embeddings are processed by $n_k$ stacked Mamba blocks (Fig.~\ref{fig:evlmamba}), where each block applies a depthwise convolution, gated activation, and selective \gls{ssm} (Section~\ref{sec:Preliminaries}) with a residual connection and Layer Normalization (LN):

\begin{equation}
\begin{aligned}
\label{eq:mamba}
\tilde{\mathbf{H}}^{k+1} &= \mathbf{H}^{k} + \mathrm{Mamba}(\mathbf{H}^{k}), \\
\mathbf{H}^{k+1} &= \mathrm{LN}\!\left(\tilde{\mathbf{H}}^{k+1}\right)
\end{aligned}
\end{equation}
where \mbox{$\mathbf{H}^{k} \in \mathbb{R}^{L \times d}$} denotes the hidden representation at the \mbox{$k$-th} block. 
Finally, the output projection layer $\phi_\text{f}$ maps the final hidden 
state $\mathbf{H}^{n_k}$ to the \gls{nig} parameters 
$(\gamma, \nu, \alpha, \beta)$ over the velocity likelihood, enabling uncertainty-aware velocity estimation as described next.


\subsubsection{\textbf{Evidential Velocity Regression}}
Velocity estimation from onboard sensors is inherently uncertain due to nonlinear vehicle dynamics and sensor noise.
To account for this uncertainty and ensure reliable velocity correction, we adopt an evidential regression framework~\cite{evidential_regression} to 
estimate both the most probable velocity and its associated uncertainty.
The true velocity $\vv_t$ is assumed to follow a Gaussian distribution with unknown mean $\mu$ and variance $\sigma^2$. 
The uncertainty is modeled following deep evidential regression~\cite{evidential_regression} by predicting the parameters of a \acrfull{nig} prior over the Gaussian mean and variance.
\begin{equation}
\begin{gathered}
\label{eq:NIG}
v_t \sim \mathcal{N}(\mu, \sigma^2), \\
\mu \sim \mathcal{N}\!\left(\gamma, \sigma^2{\nu^{-1}}\right), \sigma^2 \sim \Gamma^{-1}(\alpha, \beta), \\
p(v_t \mid \gamma, \nu, \alpha, \beta) = \mathrm{NIG}(\gamma, \nu, \alpha, \beta)
\end{gathered}
\end{equation}
where $\gamma$ represents the mean of the velocity prediction, $\nu$ is the inverse variance weight. The parameters $\alpha$ and $\beta$ define the shape and scale of the inverse gamma prior~\cite{evidential_traj}. 

The \gls{nig} distribution can be interpreted as a higher-order evidential distribution over the underlying lower-order likelihood from which observations are drawn as shown in Fig~\ref{fig:evlmamba}.
Unlike MC Dropout and ensemble-based approaches, which require multiple forward passes or models, evidential learning operates in a single forward pass. It enables the direct estimation of mean velocity and both uncertainties from the \gls{nig} parameters $m =(\gamma, \nu, \alpha, \beta)$, as follows:
\begin{equation}
\label{eq:mamba vel estimation}
\tilde{v}_t = \gamma, \quad \delta_{\tilde{v}} = 
\underbrace{\frac{\beta}{\alpha - 1}}_{\text{aleatoric}}
+
\underbrace{\frac{\beta}{\nu(\alpha - 1)}}_{\text{epistemic}}
\end{equation}
where $\delta_{\tilde{v}}$ denotes the total uncertainty in velocity estimate.

In \gls{evimamba}, the velocity $\vv_t$ is modeled independently along each axis. Therefore, the output projection layer $\phi_\text{f}$ predicts the \gls{nig} parameters $m$ of each component \mbox{$\vv_t = [v_x,v_y]^\top$} for evidential velocity regression. 
The raw network outputs after $\phi_\text{f}$ are transformed into valid \gls{nig} parameters using the softplus function to enforce $\nu,\beta > 0$, while alpha is constrained as $\alpha > 1$ for finite predictive variance. 
Finally, the training objective of \gls{evimamba} combines a \gls{nll} term for data fitting and a regularization term to penalize inaccurate uncertainty estimates, as defined below:
\begin{equation}
\begin{aligned}
\mathcal{L} &= \mathcal{L}_{\mathrm{NLL}} + \lambda \mathcal{L}_{\mathrm{reg}}, \quad
\mathcal{L}_{\mathrm{reg}} = |v_t - \gamma| (2\nu + \alpha) \\
\mathcal{L}_{\mathrm{NLL}} &=
\frac{1}{2}\log\!\left(\frac{\pi}{\nu}\right)
\!-\! \alpha \log(\Omega)\! - \!\log\!\Gamma\!\left(\alpha\! +\! 0.5 \right) \\
& \quad+\! \log\Gamma(\alpha)\!  + \left(\alpha + 0.5\right)
\log\!\left(\nu (v_t - \gamma)^2 + \Omega \right),
\end{aligned}
\end{equation}
Here $\lambda$ is a regularization weight and \mbox{$\Omega = 2\beta(1 + \nu)$}.


The proposed \gls{evimamba} architecture performs sequence-to-one velocity regression, estimating $\tilde{\vv}_t$ from $\mO_t$ via $n_k$ stacked Mamba blocks and an evidential output head. The linear-time complexity of Mamba-\gls{ssm}~\cite{mamba} combined with lightweight evidential uncertainty estimation supports real-time onboard deployment.

\subsection{ES-EKF Based IMU Mechanization}




We employ an~\gls{esekf} for precise 3D localization in GNSS-denied environments, following~\cite{es-ekf2}. 
The \gls{imu} measurements are standstill bias-corrected and transformed to the vehicle \gls{cog} prior to propagation, with input $u_t = [\va_t, \boldsymbol{\omega}_t]^\top$ where $\va_t \in \mathbb{R}^3$ and $\boldsymbol{\omega}_t \in \mathbb{R}^3$ denote linear acceleration and 
angular velocity. 
The sensor biases are initialized from standstill calibration and refined online within the \gls{esekf} as part of the state vector. 
The state vector comprises position, velocity, orientation, and sensor biases:
\begin{align}
\label{eq:nominal_state}
    \vx = [\vp,\vv,\mathbf{R}_{\text{v}}^{\text{g}},\vb_a,\vb_{\omega}]^\top
\end{align}
where $\mathbf{R}_{\text{v}}^{\text{g}} \in SO(3)$ denotes the rotation from vehicle to 
global frame.

\subsubsection{A-Priori Prediction}
The nominal state kinematics are obtained by integrating high-frequency \gls{imu} data $\vu_t$. The \mbox{\textit{a priori}} nominal state $\hat{\vx}_{t+1}^{-}$ propagation is defined as: 
\begin{equation}
\label{eq:nominal_state_prediction}
\hat{\vx}_{t+1}^{-} =
\begin{bmatrix}
\vp_t\!+\! \vv_t \Delta t\!
  +\! 0.5\left(\mathbf{R}_t(\va_t\! - \!\vb_{a,t}) \!+\! \vg\right)\Delta t^2 \\
\vv_t\! +\! \left(\mathbf{R}_t(\va_t - \vb_{a,t}) + \vg\right)\Delta t \\
\mathbf{R}_t \exp\left([(\boldsymbol{\omega}_t - \vb_{\omega,t})\Delta t]_\times\right) \\
\vb_{a,t} \\
\vb_{\omega,t}
\end{bmatrix}
\end{equation}

Here, $\vg$ denotes the gravity vector, and $\mathbf{R}_t \equiv \mathbf{R}_{\text{v}}^{\text{g}}$
is the rotation matrix defined in Eq.~\ref{eq:nominal_state}. The operator $[\cdot]_\times$ denotes the skew-symmetric operator, and to maintain numerical stability, the rotation matrix is periodically re-orthonormalized.

In the \gls{esekf}, errors arising from process noise and model 
approximations are captured in a linearized error state $\delta\vx$ defined 
around the nominal state:
\begin{equation}
\label{eq:error state}
\delta\vx = [\delta\vp,\delta\vv,\delta\boldsymbol{\theta},\delta\vb_a,\delta\vb_{\omega}]^\top
\end{equation}
where $\delta\boldsymbol{\theta} \in \mathbb{R}^3$ represents the orientation error in the Lie algebra of $SO(3)$, avoiding singularities through the small-angle approximation~\cite{es-ekf}. By linearizing around the nominal state, the error dynamics remain Gaussian and tractable, with only the error-state 
covariance propagated during prediction:
\begin{equation}
\mathbf{P}_{t+1}^{-}  = \mathbf{F}_t \mathbf{P}_t \mathbf{F}_t^\top + \mathbf{Q}_t.
\end{equation}
where $\mathbf{F}_t$ is the linearized error-state transition 
matrix, $\mathbf{G}_t$ maps process noise into the error-state 
space, and \mbox{$\mathbf{Q}_t = \mathbf{G}_t\mathbf{Q}_w\mathbf{G}_t^\top$} 
is the discrete process noise covariance. Complete derivations 
follow standard formulations~\cite{es-ekf2}.


\subsubsection{Measurement Update} 
The \gls{evimamba} velocity estimate $(\tilde{\vv}_t,\delta_{\tilde{\vv}})$ (cf. Eq.~\ref{eq:mamba vel estimation}) is incorporated as a virtual measurement to reduce position drift during~\gls{gnss} outages. Since only planar velocity is estimated, similar to external velocity sensors~\cite{kistler_vehicle_dynamics_testing}, the measurement is augmented with a zero vertical velocity constraint valid for ground vehicles. The measurement vector $\mathbf{z}_t$ and noise covariance $\mathbf{M}_s$ are:
\begin{align}
\mathbf{z}_t = [\tilde{v}_x,\, \tilde{v}_y,\, 0]^\top, \quad
\mathbf{M}_s = \mathrm{diag}(\delta_{\tilde{v}_x},\, 
       \delta_{\tilde{v}_y},\, \sigma^2_{z})
\end{align}
where $\sigma^2_{z} \ll \delta_{\tilde{v}_x}$. The corresponding measurement innovation $\mathbf{y}_t$ and Jacobian $\mathbf{H}_t \in \mathbb{R}^{3\times15}$ are:
\begin{align}
\mathbf{y}_t &= \mathbf{z}_t - \mathbf{h}(\hat{\vx}_t^{-})
= \mathbf{z}_t - \mathbf{R}_{\mathrm{g}}^{\mathrm{v}} \, \hat{\vv}_t^{-},\\
\mathbf{H}_t &=
\begin{bmatrix}
\mathbf{0}_{3\times3} &
\mathbf{R}_{g}^{v} &
-\mathbf{R}_{g}^{v}[\hat{\vv}_t^{-}]_\times &
\mathbf{0}_{3\times3} &
\mathbf{0}_{3\times3}
\end{bmatrix}
\end{align}
where $\mathbf{R}_\text{g}^\text{v} = (\mathbf{R}_\text{v}^\text{g})^\top$ rotates velocity from the global to the vehicle frame, since \gls{evimamba} can estimate velocity only in the vehicle frame from onboard sensors. 

The ES-EKF update equations to compute the posterior error state $\delta \vx_t$ and covariance $\mathbf{P}_t$ are given by:
\begin{equation}
\begin{aligned}
\mathbf{K}_t &= \mathbf{P}_t^{-} \mathbf{H}_t^\top
\left(\mathbf{H}_t \mathbf{P}_t^{-}  \mathbf{H}_t^\top + \mathbf{M}_s\right)^{-1}, \\
\delta \vx_t &= \mathbf{K}_t \mathbf{y}_t, \\
\mathbf{P}_t   &= (\mathbf{I} - \mathbf{K}_t \mathbf{H}_t)\mathbf{P}_t^{-} (\mathbf{I} - \mathbf{K}_t \mathbf{H}_t)^\top + \mathbf{K}_t \mathbf{M}_s \mathbf{K}_t^\top.
\end{aligned}
\end{equation}
Here, $\mathbf{K}_t$ denotes the Kalman gain, and $\mathbf{P}_t^{-}$ the prior error-state covariance.  Finally, the posterior error state is injected into the nominal state for correction as
\begin{equation}
\hat{\vx}_t = \hat{\vx}_t^{-} \oplus \delta \vx_t, \quad \hat{\mathbf{R}}_t = \hat{\mathbf{R}}_t^{-} \exp([\delta \boldsymbol{\theta}_t]_\times).
\end{equation}
where position, velocity, and biases are updated additively, while the orientation is corrected multiplicatively. After the update, the error state is reset to zero for the next iteration.

\section{Dataset}

Most existing navigation approaches using proprioceptive sensors focus primarily on \gls{imu}-based localization~\mbox{\cite{imu_survey,survey_localization}}. These methods are typically evaluated on the KITTI~\cite{kitti} dataset or other proprietary datasets~\cite{lstmIMU,sven}. In contrast, this work leverages both \gls{imu} and additional onboard proprioceptive sensors for localization. Since KITTI does not provide the required onboard sensor modalities, the proposed method is evaluated on the publicly available \gls{revsted}~\cite{revsted}.

The \gls{revsted} dataset comprises five hours of driving data collected using the test vehicle shown in Fig.~\ref{fig:piml_architecture} at the CARISSMA test track~\cite{carissma_outdoor_test_site}. It provides space-time synchronized measurements from a navigation-grade IMU~\cite{adma}, an external velocity sensor~\cite{kistler_vehicle_dynamics_testing}, and onboard vehicle sensors. Introduced in our prior work~\cite{abinav}, it further includes centimeter-level accurate vehicle positions obtained via \gls{ins} and RTK-based positioning, serving as ground truth for localization tasks. 
The \gls{revsted} provides onboard sensor signals $\vo_t$ at 50 Hz, defined as follows.
\begin{equation}
\label{equation:obd_data}
\vo = \begin{bmatrix}
v_\text{s}, \delta_{\text{sw}}, \dot{\psi}, a_{\text{y}}, p_{\text{br}}, v_{\text{fl}}, v_{\text{fr}}, v_{\text{rl}}, v_{\text{rr}}
\end{bmatrix}^\mathrm{T},
\end{equation}
where $v_{\text{s}}$ denotes speedometer reading, $\delta_{\text{sw}}$ the steering wheel position, $\dot{\psi}$ the yaw rate, \mbox{$v_{\text{fl}}, v_{\text{fr}}, v_{\text{rl}}, v_{\text{rr}}$} the wheel speeds, $p_{\text{br}}$ the brake pressure, and $a_{\text{y}}$ the lateral acceleration. 

\newpage
\section{Evaluation}

This section evaluates \gls{evimamba} across three aspects: 
velocity estimation accuracy, the impact of velocity 
correction on IMU-based localization, and real-time 
capability on edge hardware.

\subsection{Experimental Setup}

The proposed \gls{evimamba} is evaluated on the \gls{revsted} dataset, using a 70:10:20 split into training, validation, and test sets, as proposed in~\cite{abinav}.
The Mamba-based velocity estimation model uses $n_k=4$ stacked Mamba blocks, with \mbox{$L=250$} and \mbox{$d=256$}. It is trained using Adam (\mbox{$lr=10^{-4}$}, cosine schedule,~\mbox{$b=250$}) for 25 epochs.
The regularization weight for the evidential loss function was set to \mbox{$\lambda =4$} based on a hyperparameter study. All training was performed on a single NVIDIA Quadro RTX 5000 GPU.

The \gls{esekf} process noise parameters, including \gls{imu} noise characteristics $\sigma_a,\sigma_\omega$ and bias random walk models $\sigma_{b_a},\sigma_{b_\omega}$, are initialized from the sensor specifications~\cite{adma} and subsequently refined on the validation set to account for modeling inaccuracies. 

The accuracy of vehicle velocity estimation is evaluated using the widely adopted \gls{mae} and \gls{rmse}~\cite{abinav} metric. Vehicle localization accuracy during inertial navigation is assessed using maximum position drift~\cite{sven,wheel-imu}.

\subsection{Evaluation and Results}
\label{section:Evaluation and results}

\subsubsection{Vehicle Velocity Estimation Accuracy} 

The proposed \gls{evimamba} architecture achieves the highest velocity estimation accuracy among all evaluated methods using onboard sensor data, as shown in Table~\ref{tab:1}. The \gls{osd}-baseline represents a kinematic model-based velocity estimation derived from wheel speed measurements~\cite{abinav}. The transformer-based architecture employs an encoder to quantify vehicle velocity using the same input as \gls{evimamba}.
Our proposed method outperforms the transformer by approximately
$30\%$ in lateral velocity estimation, demonstrating the advantage of the Mamba-\gls{ssm} for sequential velocity estimation.

\begin{figure}[htbp!]
    \centering
        \includegraphics[scale=0.75, trim=5pt 0pt 0pt 0pt, clip]{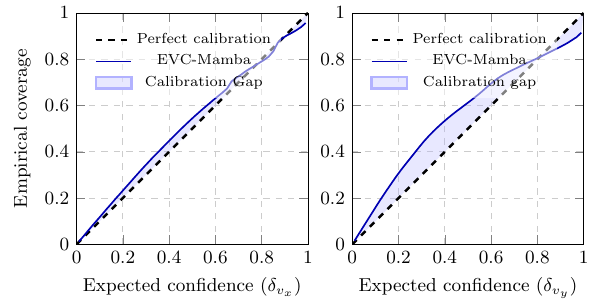}
    \caption{Uncertainty calibration plots for predicted uncertainties 
    $\delta_{\tilde{v}_x}$ and $\delta_{\tilde{v}_y}$ of \gls{evimamba}.}
    \label{fig:uncertainty calibration}
\end{figure}

The reliability of the uncertainty estimates is validated using calibration plots in Fig.~\ref{fig:uncertainty calibration}. A perfectly calibrated model follows the diagonal, where empirical coverage matches the expected confidence level, while deviations above and below indicate underconfidence and overconfidence, respectively. 

As shown, $\delta_{\tilde{v}_x}$ closely follows the diagonal, indicating well-calibrated uncertainty estimates. In contrast, $\delta_{\tilde{v}_y}$  exhibits mild underconfidence, which may be partly attributed to the predominance of near-zero lateral velocities in driving scenarios. This is further supported by Pearson correlation coefficients of $r = [0.47, 0.45]$ with $t^* = [227, 219]$ for $\delta_{\tilde{v}_x}$ and $\delta_{\tilde{v}_y}$ respectively, confirming that the predicted uncertainty meaningfully tracks velocity prediction error.


As the velocity estimates from \gls{evimamba} demonstrate good accuracy and 
calibrated uncertainty, they can serve as a reliable virtual 
corrections during \gls{gnss}-denied conditions.

\begin{table}[htbp]
\centering
\caption{Comparison of vehicle velocity estimation on \mbox{\gls{revsted}~\cite{revsted}} using \gls{mae} and \gls{rmse} metrics.}
\begin{tabular}{lcccc}
\toprule 
\textbf{Models} & \multicolumn{2}{c}{$v_x$ (m/s)} & \multicolumn{2}{c}{$v_y$ (m/s)} \\ 
\cmidrule(lr){2-3} \cmidrule(lr){4-5}
 & \gls{mae} & \gls{rmse} & \gls{mae} & \gls{rmse} \\ 
\midrule
OSD-Baseline~\cite{abinav} & 0.18 & 0.21 & 0.048 & 0.081 \\
Transformer~\cite{attention}   & 0.03 & 0.08 & 0.008 & 0.017 \\
\gls{evimamba} (Ours) & \textbf{0.03} & \textbf{0.07} & \textbf{0.006} & \textbf{0.011} \\
\bottomrule
\end{tabular}
\label{tab:1}
\end{table}

\subsubsection{Accuracy of Inertial Navigation with Velocity Correction} 

The \gls{imu} localization performance during GNSS outages was evaluated across three outage durations, each tested in five non-overlapping scenarios. Results are summarized in Table~\ref{tab:localization_results}.
The No Correction baseline represents pure \gls{imu} dead reckoning 
using \gls{esekf} without velocity correction, while all other methods 
incorporate their respective velocity estimates as virtual measurements.
Due to unbounded error growth, maximum position drift scales nearly quadratically in the No correction baseline.
The introduction of velocity correction substantially reduces drift; even the coarse velocity estimate from the \gls{osd}-Baseline~\cite{abinav} cuts maximum drift by roughly 80\%, constraining its growth to near-linear behavior.
Learning-based velocity estimation extends this advantage further, with \gls{evimamba} achieving the lowest drift among all onboard sensor-based velocity correction methods.

\begin{figure}[tbhp!]
    \centering
        \includegraphics[scale=0.8, trim=0pt 0pt 0pt 0pt, clip]{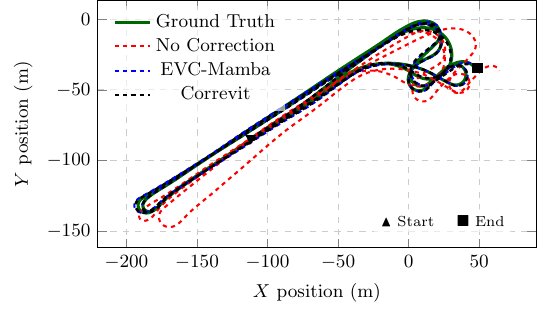}
    \caption{Qualitative vehicle localization performance during a 5-min \gls{gnss} outage in \gls{revsted}~\cite{revsted}.}
    \label{fig:trajectory}
    \vspace{-1.5em}
\end{figure}

Notably, the external velocity sensor (Correvit) serves as the practical upper bound, representing the \gls{imu} localization achievable with a dedicated hardware correction. The proposed uncertainty-aware velocity correction reaches within roughly 10\% of this upper bound across all evaluated  durations, as shown in Table~\ref{tab:localization_results}. A representative trajectory for a 5-minute outage is shown in 
Fig.~\ref{fig:trajectory}. 
This highlights \gls{evimamba} as a cost-efficient, hardware-free 
alternative for reliable localization correction during \gls{gnss} outages.

\begin{table}[htbp!]
\centering
\caption{Maximum position drift (m) during inertial navigation across GNSS outage durations.}
\label{tab:localization_results}
\begin{tabular}{lccc}
\toprule
 & \multicolumn{3}{c}{\textbf{GNSS Outage Duration}} \\
\cmidrule(lr){2-4}
\textbf{Method} & \textbf{2 min} & \textbf{5 min} & \textbf{10 min} \\
\midrule
No Correction                & $7.8 \pm 2.3$ & $31.8 \pm 5.1$  & $142 \pm 29.9$ \\
OSD-Baseline~\cite{abinav}   & $1.6 \pm 0.4$ & $4.8 \pm 2.1$   & $13.0 \pm 3.4$   \\
Transformer~\cite{attention} & $1.2 \pm 0.5$ & $2.9 \pm 1.5$   & $7.2 \pm 2.6$   \\
\gls{evimamba} (Ours)      & \textbf{1.0 $\pm$ 0.5} & \textbf{2.4 $\pm$ 1.3} & \textbf{5.9 $\pm$ 2.0} \\
Correvit~\cite{kistler_vehicle_dynamics_testing}   & \textit{0.9 $\pm$ 0.5} & \textit{2.2 $\pm$ 1.2} & \textit{5.4 $\pm$ 1.7} \\
\bottomrule
\multicolumn{4}{l}{\scriptsize \textbf{Bold}: Best learning-based velocity correction; 
\textit{Italic}: Hardware-based upper bound.}
\end{tabular}
\end{table}

\subsubsection{Real-Time Capability}
The \gls{evimamba} based velocity estimation was evaluated for real-time deployment on an NVIDIA Orin edge device. It achieves an inference latency of 20-24 ms and can operate at $40$ Hz. With only 22.8 MFLOPs per inference pass, \gls{evimamba} is roughly 3 times computationally more efficient than the Transformer baseline. Despite this reduced complexity, it maintains superior velocity estimation, making it suitable for edge deployment.

\section{Conclusion}

This work presents \gls{evimamba}, a learning-based architecture providing uncertainty-aware velocity correction for inertial navigation in \gls{gnss}-denied environments. By leveraging onboard sensors such as steering angle and wheel speeds, the proposed approach constructs a precise virtual velocity sensor, eliminating reliance on dedicated external hardware. A Mamba-based \gls{ssm} captures temporal vehicle dynamics efficiently, while evidential deep learning with a Normal-Inverse-Gamma distribution provides principled, single-pass uncertainty quantification. The resulting velocity estimate and its uncertainty are integrated into an \gls{esekf} as virtual measurements, constraining velocity and reducing position drift during prolonged \gls{gnss} outages. Experimental evaluation demonstrates that \gls{evimamba} achieves localization accuracy within 10\% of a dedicated Correvit-class hardware sensor, while outperforming all onboard-sensor-based baselines. 
\gls{evimamba} operates in real-time, offering a cost-efficient, external hardware-free alternative for proprioceptive localization.

Future work will explore transfer learning across vehicle platforms and online model adaptation during periods of satellite availability.

\section{Acknowledgement}
This work was funded by the Deutsche Forschungsgemeinschaft (DFG, German Research Foundation) – FIP 135/1, Project Number 549102058, and by the Federal Ministry of Education and Research of Germany (BMBF) – HyMne2, Project Number 13FH7I13IA.






\bibliographystyle{IEEEtran}
\bibliography{root}

\end{document}